\documentclass[conference]{IEEEtran}
\IEEEoverridecommandlockouts
\usepackage[numbers]{natbib}
\usepackage{amsmath,amssymb,amsfonts}
\usepackage{algorithmic}
\usepackage{graphicx}
\usepackage{textcomp}
\usepackage{listings}
\usepackage[dvipsnames,table,xcdraw]{xcolor}
\usepackage{pdfpages}
\usepackage{booktabs}
\usepackage[utf8]{inputenc}

\def\BibTeX{{\rm B\kern-.05em{\sc i\kern-.025em b}\kern-.08em
    T\kern-.1667em\lower.7ex\hbox{E}\kern-.125emX}}

\begin{document}

\title{Gradient-based Fuzzy System Optimisation via Automatic Differentiation -- FuzzyR as a Use Case\\
}

\author{\IEEEauthorblockN{Chao Chen}
\IEEEauthorblockA{\textit{School of Computer Science} \\
\textit{University of Nottingham}\\
Nottingham, UK \\
chao.chen@nottingham.ac.uk}
\and
\IEEEauthorblockN{Christian Wagner}
\IEEEauthorblockA{\textit{School of Computer Science} \\
\textit{University of Nottingham}\\
Nottingham, UK \\
christian.wagner@nottingham.ac.uk}
\and
\IEEEauthorblockN{Jonathan M. Garibaldi}
\IEEEauthorblockA{\textit{School of Computer Science} \\
\textit{University of Nottingham}\\
Nottingham, UK and Ningbo, China \\
jon.garibaldi@nottingham.edu.cn}
}

\maketitle

\begin{abstract}

Since their introduction, fuzzy sets and systems have become an important area of research known for its versatility in modelling, knowledge representation and reasoning, and increasingly its potential within the context explainable AI.
While the applications of fuzzy systems are diverse, there has been comparatively little advancement in their design from a machine learning perspective. 
In other words, while representations such as neural networks have benefited from a boom in learning capability driven by an increase in computational performance in combination with advances in their training mechanisms and available tool, in particular gradient descent, the impact on fuzzy system design has been limited.
In this paper, we discuss gradient-descent-based optimisation of fuzzy systems, focussing in particular on automatic differentiation--crucial to neural network learning--with a view to free fuzzy system designers from intricate derivative computations, allowing for more focus on the functional and explainability aspects of their design. 
As a starting point, we present a use case in FuzzyR which demonstrates how current fuzzy inference system implementations can be adjusted to leverage powerful features of automatic differentiation tools sets, discussing its potential for the future of fuzzy system design.

\end{abstract}


\section{Introduction}
\label{sec:intro}
Since the introduction of fuzzy sets and systems by Zadeh in 1965~\cite{Zadeh1965}, this domain has gained substantial traction, evolving into a prominent field of research and a versatile tool for modelling, control and reasoning 
 across various disciplines~\cite{Karnik2001, Mendel2002, Mendel2006, Castillo2012, Soria2013}. 
Over the past few years, a variety of toolkits have emerged in different programming languages that significantly aid in the dissemination, accessibility and practical implementation of theoretical and applied research in fuzzy systems~\cite{Alcala2016, Pontes2023}. 
For example, there are MATLAB\textsuperscript{\textregistered} toolboxes~\citep{Babuska2000, Castro2007, FuzzyMathWorks}; GUAJE, FisPro, Juzzy and Juzzy Online for Java~\citep{Alonso2011, Guillaume2012, Wagner2013, Wagner2014}, and Fuzzycreator and Scikit-Fuzzy for Python~\citep{McCulloch2017, Warner2019}. 
In previous work, we introduced FuzzyR which is an open source toolbox in the R programming language for modelling different types of fuzzy inference system, including type-1, interval type-2, hierarchical, and non-singleton models~\cite{Chen2020FuzzyR, Chen2021FuzzyR, Razak2021FuzzyR}.

Despite significant advancements in the domain of fuzzy systems, a critical and broader gap is evident in the field: the limited focus on optimisation within the research area, while preserving the core attributes of explainability and interpretability.
Various fuzzy models have been proposed that showcase these intrinsic advantages, but often the performance optimisation aspect is not as emphasised~\cite{Chen2021FuzzyR, Razak2021FuzzyR, Wang2023Constrained}. 
This gap is largely attributable to the lack of integrated optimisation capabilities within the toolkits. 
On the other hand, while the exploration of gradient-based methods has been fairly extensive since the 1990s, the choices of modelling configurations, including the selection of membership functions, have normally been limited mainly due to the challenges associated with derivative computations~\cite{Jang1993, Wu2012MFs, Wu2020Optimise}. 
Consequently, the use of gradient-based methods was discouraged, especially for type-2 modelling, which may require more intricate derivative calculations~\cite{Mendel2004Derivatives, Mendel2014}.
Hence, even in the cases where optimisation methods are provided, the options are usually limited to methods such as genetic algorithms, particle swarm optimisation, and simulated annealing~\cite{FuzzyMathWorks, Wagner2010a, Melin2013, Wu2014, Almaraashi2016}. 

In contrast, deep learning has achieved remarkable success with the implementation of gradient-based parameter learning methods~\cite{Lecun1998, AlexNet2012, Lecun2015}. 
The effectiveness of state-of-the-art deep learning models lies in their efficiency in navigating large parameter spaces based on gradient-based methods, crucial for complex systems~\cite{Goodfellow2016}. 
We note that within machine learning, tool development has focused strongly on gradient-based optimisation, as probably most famously represented by the autograd feature~\cite{Pytorch2017Autograd}, a cornerstone of deep learning frameworks. 
The latter significantly eases the differential process by automating derivative computations, an integral aspect of deep learning success in complex model designs.

One of the most commonly used machine learning frameworks which includes the `autograd' feature is PyTorch~\cite{Paszke2019PyTorch}.
Although PyTorch is a Python library, a Torch for R package was recently released that allows PyTorch-like functionalities to be used in the R programming language~\cite{Keydana2020, Falbel2023}.
This makes it possible for FuzzyR to benefit from the autograd feature for easier optimisation.
The autograd feature liberates designers from the intricacies of derivative calculations, allowing them to focus more on the creative aspects of fuzzy system design, such as architecture innovation and the exploration of alternative functions and operators.
This paper introduces the integration of automatic differentiation with FuzzyR as a use case.
Such integration provides the potential for more flexible and more advanced fuzzy inference system design, including support for a broader range of membership functions and fuzzy operators during inference processes such as fuzzification and deffuzification.

The main aim of this paper is to introduce the autograd-enabled FuzzyR toolbox with examples to illustrate how it may be used for optimisation when building a fuzzy model.
We proceed to discuss the potential of leveraging some of the great advances 
in deep learning, in order to develop more advanced 
fuzzy systems, while preserving their unique potential for explainability.

\section{Automatic Differentiation}
This section demonstrates a simple autograd example adapted from Torch for R~\cite{Falbel2023}.
By a simple linear model with autograd capabilities, this example aims to underscore how the complexities of derivative calculations, a cornerstone in optimisation algorithms like gradient descent, can be efficiently managed.

\begin{lstlisting}[language=R, caption=Example code snippet of autograd in R]
    require(torch)
    x <- torch_tensor(3)
    w <- torch_tensor(2, requires_grad = TRUE)
    b <- torch_tensor(1, requires_grad = TRUE)
    y <- w * x + b
    y$backward()
    w$grad
    #> torch_tensor
    #>  3
    #> [ CPUFloatType{1} ]
    b$grad
    #> torch_tensor
    #>  1
    #> [ CPUFloatType{1} ]
\end{lstlisting}

In the above example, $y$ can be considered as the output of a linear model where $x$ is the input, $w$ is the weight, and $b$ is the bias.
In machine learning, the values of $w$ and $b$ are usually learnt from the data.
When the gradient descent method is used, the partial derivatives of $y$ with respect to $w$ and $b$ are calculated--a challenging process if done manually.

However, with the automatic differentiation feature, as illustrated above, there is no need to perform such calculations manually.
In the context of this example, the key lies in declaring $w$ and $b$ as tensors that require a gradient. 
The tensor, in high-level terms, can be thought of as a multidimensional data structure, which is fundamental to operations in machine learning and deep learning frameworks. 
When a tensor is declared with the `requires\_grad' attribute set to true, as in the case of $w$ and $b$, it signals to the Torch framework that every operation applied to these tensors should be tracked for the automatic differentiation process. 
Thus, tensors are not just containers for numerical data, but are dynamic structures that facilitate the calculation and propagation of gradients in computational graphs.

In the example, after operations ($y = w * x + b$) in the forward pass, the $backward()$ method can be used to automatically compute the gradient of $y$ with respect to the tensors $w$ and $b$.
Then, $w\$grad$ and $b\$grad$ can be used to access relevant gradients.

As discussed above, to use autograd, the data flow and operations need to be based on tensors. 
While many built-in functions and operators in R have been supported by Torch for R, customised functions, especially, often require modifications to be compatible with autograd.

A case in point is the commonly used \emph{apply} function in R, which is very useful for applying a function to the margins of an array or matrix. 
However, this function is currently not supported by Torch for R. 
To illustrate this limitation, consider the following example where the apply function is used:
\begin{lstlisting}[language=R, caption=Example code to show that the $apply$ funtion is not supported]
    require(torch)
    x <- torch_tensor(matrix(rnorm(10), nrow=2))
    w <- torch_tensor(matrix(rnorm(10), nrow=2), requires_grad = T)
    x * w
    #> torch_tensor
    #> 1.4192  0.0270 -0.1124  0.2605  0.0537
    #> 1.1509 -0.3120  0.0478 -0.5553  0.0104
    #> [ CPUFloatType{2,5} ][ grad_fn = <MulBackward0> ]
    y <- apply(x * w, 1, sum)
    y
    #> [1] 1.6479214 0.3417696
    y[1]$backward()
    #> Error in y[1]$backward : $ operator is invalid for atomic vectors
\end{lstlisting}

In this example, the $apply$ function is used to calculate the sum of the rows for the matrix $x * w$. 
The resulting object from apply is not a tensor, hence autograd is not supported. 
This necessitates an alternative approach to utilise autograd. 
An example solution, which achieves a similar outcome but in a manner compatible with autograd, is provided below:
\begin{lstlisting}[language=R, caption=Example solution using torch operations for autograd compatibility]
    require(torch)
    x <- torch_tensor(matrix(rnorm(10), nrow=2))
    w <- torch_tensor(matrix(rnorm(10), nrow=2), requires_grad = T)
    x * w
    #> torch_tensor
    #> 1.4192  0.0270 -0.1124  0.2605  0.0537
    #> 1.1509 -0.3120  0.0478 -0.5553  0.0104
    #> [ CPUFloatType{2,5} ][ grad_fn = <MulBackward0> ]
    y <- torch_stack(
            lapply(torch_unbind(x * w, dim = 1), sum))
    y
    #> torch_tensor
    #> 1.6479
    #> 0.3418
    #> [ CPUFloatType{2} ][ grad_fn = <StackBackward0> ]
    y[1]$backward(retain_graph = T)
    y[2]$backward()
    w$grad
\end{lstlisting}

This solution demonstrates the use of $torch\_stack$ and $lapply$ along with $torch\_unbind$ to perform operations analogous to $apply$, but in a way that maintains compatibility with the autograd system.

It should be mentioned that this paper does not aim to provide an exhaustive list of functions supported or unsupported by Torch for R in the context of autograd compatibility. 
The scope of such an endeavour is vast, and the functionalities within Torch for R and PyTorch are continually evolving. 
Therefore, users are encouraged to refer to the official manual for the most up-to-date and comprehensive information on function support and compatibility.
In addition, other related resources, such as community forums, can be accessed for the latest developments, user experiences, and a wider range of examples and use cases.


\section{Modifications in FuzzyR for~Automatic~Differentiation}
Making FuzzyR fully compatible with autograd requires significant modifications. 
This section outlines some initial changes that have been implemented to adapt FuzzyR for this purpose. 
Specifically, we focus on modifications that render a demonstrative Mamdani-type fuzzy inference system compatible with autograd.

These initial changes are the groundwork for further integration and optimisation within FuzzyR, utilising the advanced capabilities of autograd. 
Note that modularity and layer-based structure are emblematic of deep learning architectures. 
In deep learning, the ability to construct complex models using multiple building blocks (such as layers and modules) has been a paradigm shift, allowing the creation of highly adaptable and sophisticated systems. 
Similarly, the integration of automatic differentiation capabilities into FuzzyR unlocks a transformative level of flexibility.
This advancement may significantly shift the focus toward more strategic aspects of fuzzy system development, empowering designers to construct fuzzy systems in a more modular fashion (e.g., something like hierarchical fuzzy systems~\cite{Razak2021FuzzyR}).
Designers will be able to concentrate on the conceptual and architectural design of these systems, rather than being bogged down by the manual calculations of derivatives for optimisation.
On the other hand, the adaptations and methodologies described here can serve as references for extending the autograd compatibilities of FuzzyR (or other toolkits) in the future.


Specific changes are introduced in the following.
Membership functions are fundamental components of fuzzy logic systems.
The trapezoidal membership function is one of the commonly used types of membership function.
The following code is the original implementation of this function in FuzzyR.

\begin{lstlisting}[language=R, caption=Original implementation of the trapezoidal membership function]
    trapmf <- function(mf.params) {
        a <- mf.params[1]
        b <- mf.params[2]
        c <- mf.params[3]
        d <- mf.params[4]

        if (length(mf.params) == 5) {
            h <- mf.params[5]
        } else {
            h <- 1
        }

        trapmf <- function(x) {
            y <- pmax(pmin((x - a) / (b - a), h, (d - x) / (d - c)), 0)
            y[is.na(y)] <- h
            y
        }
    }
\end{lstlisting}

However, this original implementation requires the use of functions $pmax$ and $pmin$, which are not supported by Torch for R. 
To make the trapezoidal membership function compatible with autograd in Torch for R, an alternative approach is necessary.
The revised implementation, presented below, addresses compatibility issues, ensuring that the function can operate with torch tensors and is suitable for autograd~processing.

\begin{lstlisting}[language=R, caption=Autograd compatible implementation of the trapezoidal membership function]
    trapmf.torch <- function(mf.params) {
        a <- mf.params[1]
        b <- mf.params[2]
        c <- mf.params[3]
        d <- mf.params[4]

        if (length(mf.params) == 5) {
            h <- mf.params[5]
        } else {
            h <- 1
        }

        trapmf.torch <- function(x) {
            x <- torch_tensor(x)
            y <- torch_zeros(length(x))

            # Check conditions and calculate values accordingly
            mask1 <- (x > a) & (x < b)
            mask2 <- (x >= b) & (x <= c)
            mask3 <- (x > c) & (x < d)
            y[mask1] <- torch_minimum((x[mask1] - a) / (b - a), h)
            y[mask2] <- h
            y[mask3] <- torch_minimum((d - x[mask3]) / (d - c), h)

            y
        }
    }
\end{lstlisting}

The $evalmf$ function plays a crucial role in evaluating membership grades for given crisp inputs $x$. 
In its original form, this function may have been limited to returning crisp values only. 
To fully leverage the capabilities of autograd in Torch for R, modifications are necessary to ensure that $evalmf$ returns torch tensors instead.
The code below presents the autograd-compatible implementation of the evalmf function. 
This revised implementation ensures that the function can handle torch tensors, which is a critical requirement for autograd processing.
However, it should be noted that further modifications may be required for the $evalmf$ function to fully support all types of membership functions, including non-singleton types.

\begin{lstlisting}[language=R, caption=Autograd compatible implementation of the evalmf function]
    evalmf <- function(...) {
        params <- list(...)
        params.len <- length(params)

        x <- params[[1]]

        if (params.len == 3) {
            MF <- genmf(mf.type = params[[2]], mf.params = params[[3]])
        } else if (params.len == 2) {
            MF <- params[[2]]
        } else if (params.len == 6 || params.len == 7) {
            # TODO: autograd for non-singleton.
            return(evalmf.ns(...))
        } else {
            stop("incorrect parameters")
        }

        y <- sapply(c(MF), function(F) F(x))
        if (is.list(y)) {
            y <- y[[1]]
        }
        if (!is(y, "torch_tensor")) {
            y <- torch_tensor(y)
        }
        if (length(dim(y)) > 1) {
            y <- torch_squeeze(y, 2)
        }
        y
    }
\end{lstlisting}

In addition to the modifications made to the evalmf function, several other key functions within FuzzyR have been updated to enhance compatibility with autograd. 
These functions include $gensurf$, $plotvar$, $evalfis$, $defuzz$, among others. 
Each of these functions plays a crucial role in fuzzy inference system modelling, and their adaptation is essential for leveraging the full capabilities of autograd in Torch for R.
The details of these changes are not extensively covered in this paper, but can be found in the source files of the FuzzyR package on CRAN~\cite{FuzzyR}. 

\section{Experimental Demonstration}
In this section, the focus is on demonstrating the practical application of automatic differentiation in the context of an enhanced FuzzyR framework. 
Specifically, we optimise a Mamdani-type fuzzy inference model, using the well-known IRIS classification problem as a test case. 
This serves as a tangible example of how the integration of autograd into FuzzyR can be utilised in real-world scenarios.

This demonstration is primarily aimed at showcasing the ease with which gradient-based optimisation can be implemented once a fuzzy system is built, particularly when enhanced with autograd capabilities as in FuzzyR. 
Although this experiment does not directly illustrate the development of more flexible or complex module-based fuzzy system modelling, it sets the groundwork for such advancements. 
By integrating autograd, it opens up opportunities for experimenting with different components of fuzzy systems, such as various types of membership function and operators, without the daunting task of manual differentiation for optimisation purposes.
For example, current ANFIS modelling in FuzzyR only supports generalised bell-shaped membership functions and singleton fuzzification.
This is mainly due to the complexity of manual derivative calculations.
However, with automatic differentiation, it would be much easier to extend the support of other types of membership function and non-singleton fuzzification.



\subsection{Dataset}
The Iris dataset, consisting of 150 instances, is a classic dataset in the field of machine learning~\cite{iris}. 
This dataset is renowned for its clarity and simplicity, making it an ideal choice for testing and validating machine learning models.
As illustrated in Table~\ref{tab:iris}, the Iris dataset comprises four attributes: sepal length, sepal width, petal length, and petal width. 
It classifies instances into three species of iris flowers, which are Setosa, Versicolor, and Virginica. 
For the purposes of this demonstration, all 150 instances will be utilised both as the training and the testing sets.

\begin{table}[!ht]

\caption{Data samples of the Iris flower data set.}
\begin{center}
\resizebox{.45\textwidth}{!} {

\begin{tabular}{cccccccccc}
\toprule
\bfseries Sepal Length & \bfseries Sepal Width  & \bfseries Petal Length & \bfseries Petal width & \bfseries Species  \\ 
\midrule
	5.1 & 3.5 & 1.4 & 0.2 & setosa \\
	4.9	& 3.0 & 1.4	& 0.2 & setosa \\
	7.0 & 3.2 & 4.7 & 1.4 & versicolor \\
	6.3 & 3.3 & 6.0 & 2.5 & virginica \\
\bottomrule 
\end{tabular}
}
\end{center}

\label{tab:iris}
\end{table}

\subsection{Membership Function and The Rule Set}
When constructing the fuzzy rule base, only Petal Length ($x_1$) and Petal Width ($x_2$) were used as input.
Species ($y$) is the output.
There are three trapezoidal membership functions ($Low$, $Mid$, $High$) for each input and output.
Note that for the output, $Low$, $Mid$, $High$ are used to represent species setosa, versicolor, and viginica, respectively.
The five rules used are presented below.
\begin{eqnarray*}
&\text{\textbf{IF} $x_1$ is $Low$ \textbf{AND} $x_2$ is $Low$, \textbf{THEN} y is $Low$} \\
&\text{\textbf{IF} $x_1$ is $Mid$ \textbf{AND} $x_2$ is $Mid$, \textbf{THEN} y is $Mid$} \\
&\text{\textbf{IF} $x_1$ is $High$ \textbf{AND} $x_2$ is $High$, \textbf{THEN} y is $High$} \\
&\text{\textbf{IF} $x_1$ is $Mid$ \textbf{AND} $x_2$ is $High$, \textbf{THEN} y is $High$} \\
&\text{\textbf{IF} $x_1$ is $High$ \textbf{AND} $x_2$ is $Mid$, \textbf{THEN} y is $High$} 
\end{eqnarray*}

\subsection{Code Demo}

In the following section, we delve into the main implementation details of building a Mamdani-type inference model using FuzzyR, with an emphasis on ensuring compatibility with autograd. 
The code snippet below illustrates the process of building the fuzzy inference system, specifically tailored for the Iris classification task.


\begin{lstlisting}[language=R, caption=Code snippet for building the Mamdani-type inference model]
    fis_iris <- function(theta1, theta2) {
        fis <- newfis("Iris Classification", andMethod = "prod")
        fis <- addvar(fis, "input", "Petal.Length", c(0, 1))
        fis <- addvar(fis, "input", "Petal.Width", c(0, 1))
        fis <- addvar(fis, "output", "Species", c(0.5, 3.5))
    
        fis <- addmf(fis, "input", 1, "Low", 
                    "trapmf.torch", torch_cat(list(0, 0, theta1[1:2]), dim = 1))
        fis <- addmf(fis, "input", 1, "Mid", 
                    "trapmf.torch", theta1[3:6])
        fis <- addmf(fis, "input", 1, "High", 
                    "trapmf.torch", torch_cat(list(theta1[7:8], 1, 1), dim = 1))
    
        fis <- addmf(fis, "input", 2, "Low", 
                    "trapmf.torch", torch_cat(list(0, 0, theta2[1:2]), dim = 1))
        fis <- addmf(fis, "input", 2, "Mid", 
                    "trapmf.torch", theta2[3:6])
        fis <- addmf(fis, "input", 2, "High", 
                    "trapmf.torch", torch_cat(list(theta2[7:8], 1, 1), dim = 1))
    
        fis <- addmf(fis, "output", 1, "setosa", "trapmf.torch", c(0.5, 0.5, 0.5, 2, 1))
        fis <- addmf(fis, "output", 1, "versicolor", "trapmf.torch", c(0.5, 2, 2, 3.5, 1))
        fis <- addmf(fis, "output", 1, "virginica", "trapmf.torch", c(2, 3.5, 3.5, 3.5, 1))

        # IF Petal.Length is Low and Petal.Width is Low THEN Species is setosa
        # IF Petal.Length is Mid and Petal.Width is Mid THEN Species is versicolor
        # IF Petal.Length is High and Petal.Width is High THEN Species is virginica
        # IF Petal.Length is Mid and Petal.Width is High THEN Species is virginica
        # IF Petal.Length is High and Petal.Width is Mid THEN Species is virginica
        rl <- rbind(c(1, 1, 1, 1, 1), c(2, 2, 2, 2, 1), c(3, 3, 3, 3, 1), c(2, 3, 3, 3, 1), c(3, 2, 3, 3, 1))
        fis <- addrule(fis, rl)
    }
\end{lstlisting}

The above code snippet demonstrates the use of autograd compatible membership functions, specifically $trapmf.torch$, in the construction of a fuzzy inference system. 
Notably, the syntax remains consistent with the use of FuzzyR in non-autograd contexts. 
The primary requirement for autograd compatibility is ensuring that membership functions are compatible with autograd and that the parameters for these functions are specified as tensors. 
This is essential if automatic differentiation is needed for the optimisation process.

The following code snippets detail the steps involved in data manipulation, parameter initialisation, and optimisation for the fuzzy inference model, highlighting the application of automatic differentiation in the process.

First, we focus on data preprocessing and model initialisation:
\begin{lstlisting}[language=R, caption=Code snippet for data preprocessing and model initialisation, label={lst:initialisation}]
    ## Data Preprocessing
    data.all <- iris[, 3:5]
    species <- as.factor(data.all[, 3])
    data.all[, 3] <- as.numeric(species)
    data.all <- as.matrix(data.all)
    require(caret)
    data.all[, 1:2] <- predict(preProcess(data.all[, 1:2], method = c("range")), data.all[, 1:2])
    input_stack <- data.all[, 1:2]

    ## Parameter Initialisation
    theta1 <- c(0.1, 0.39, 0.11, 0.4, 0.6, 0.89, 0.61, 0.9)
    theta2 <- c(0.1, 0.39, 0.11, 0.4, 0.6, 0.89, 0.61, 0.9)
    
    psi1 <- torch_tensor(getPSI(theta1), requires_grad = T)
    psi2 <- torch_tensor(getPSI(theta2), requires_grad = T)
    theta1 <- getTheta(psi1)
    theta2 <- getTheta(psi2)
    
    fis.init <- fis_iris(theta1, theta2)
    fis <- fis.init
\end{lstlisting}

Next, we proceed to the optimisation process. 
Parameters $theta1$ and $theta2$ were used to define input trapezoidal membership functions. 
However, during training epochs, updates to these parameters are mediated through intermediate parameters $psi1$ and $psi2$. 
This approach ensures that the membership functions adhere to specific constraints, which are essential to maintain the validity of the model, but are not the focus of this paper.
Hence, details about the constraints will not be discussed here, but will be thoroughly investigated in another work.
The following code illustrates the optimisation process, leveraging automatic differentiation:

\begin{lstlisting}[language=R, caption=Code snippet for the optimisation process based on automatic differentiation, label={lst:optimisation}]
    epochs <- 100
    stepsize <- 0.3
    err.min <- Inf
    err.all <- NULL
    theta1.all <- t(as.matrix(theta1))
    theta2.all <- t(as.matrix(theta2))

    for (i in 1:epochs) {
        y <- evalfis(input_stack, fis)
        y <- torch_squeeze(y, 2)
        err <- rmse(data.all[, 3], y)
        cat(as.numeric(err), ",")
        if (i %% 5 == 0) {
            cat("\n")
        }

        err.all <- c(err.all, as.numeric(err))
        if (err.min > as.numeric(err)) {
            err.min <- as.numeric(err)
            fis.final <- fis
        }

        if (i == epochs) {
            break
        }

        err$backward()
        psi1 <- torch_tensor(psi1 - stepsize * psi1$grad, requires_grad = T)
        psi2 <- torch_tensor(psi2 - stepsize * psi2$grad, requires_grad = T)
        theta1 <- getTheta(psi1)
        theta2 <- getTheta(psi2)
        fis <- fis_iris(theta1, theta2)

        theta1.all <- rbind(theta1.all, t(as.matrix(theta1)))
        theta2.all <- rbind(theta2.all, t(as.matrix(theta2)))
    }
\end{lstlisting}

\subsection{Results}
The results obtained are similar to the previous work~\cite{Chen2016}.
During 100 training epochs, we observed a marked decrease in root mean square error (RMSE), from 0.2872 to 0.1421, along with a reduction in misclassified cases from 22 to~4. 

\begin{figure}[!ht]
    \begin{center}
        \includegraphics[trim=0 0 0 0, clip, width=.95\linewidth]{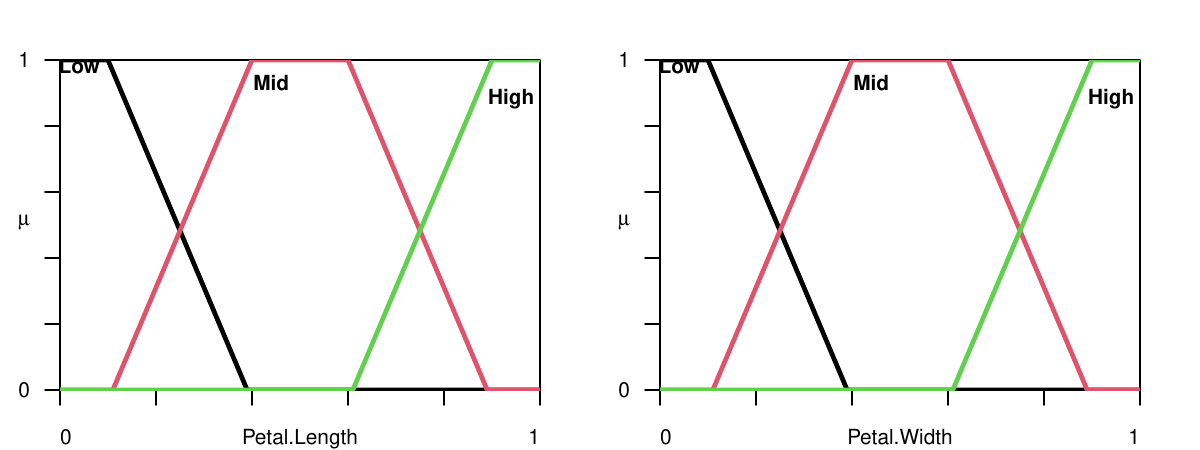}
        \caption{Initial membership functions of the inputs}
        \label{fig:InitMFs}
    \end{center}
\end{figure}

\begin{figure}[!ht]
    \begin{center}
        \includegraphics[trim=0 0 0 0, clip, width=.95\linewidth]{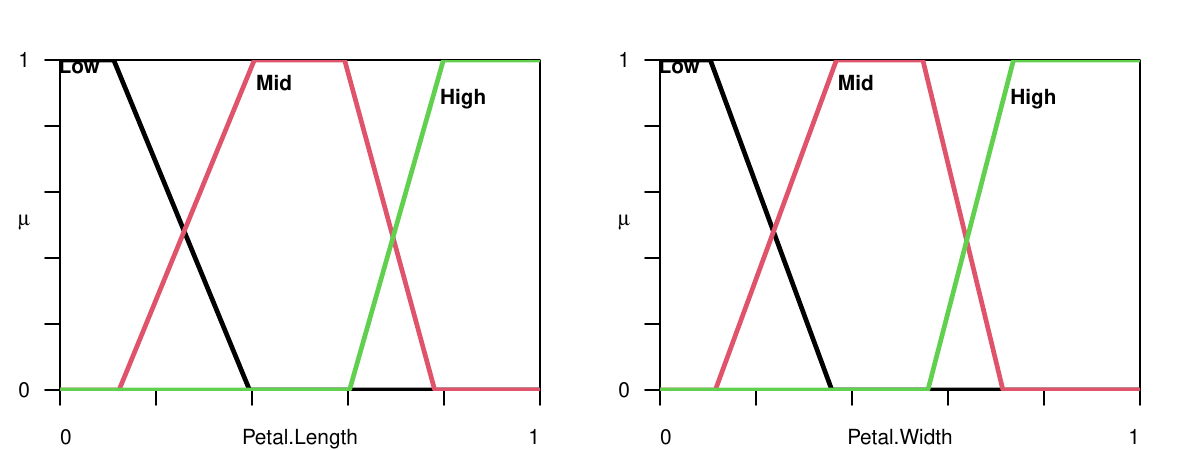}
        \caption{Final membership functions of the inputs after optimisation}
        \label{fig:FinalMFs}
    \end{center}
\end{figure}

The membership functions of the inputs before and after optimisation are presented in Figures~\ref{fig:InitMFs}~and~\ref{fig:FinalMFs} respectively.
Initially, the membership functions (`Low', `Mid', and `High') are uniformly distributed across the input range for both the petal length and the petal width, as depicted in Fig.~\ref{fig:InitMFs}. 
This evenly spaced initialisation is a standard practice in the preliminary design of fuzzy systems, serving as a neutral starting point that does not presume any data-specific biases.
However, such uniform distributions may not accurately capture the nuances of the actual data distributions present in the Iris dataset. 
The optimisation process, as illustrated in Fig.~~\ref{fig:FinalMFs}, fine-tunes these membership functions, resulting in a more data-centric alignment. 
Consequently, the final membership functions retain their interpretability--a quintessential feature of fuzzy systems--while also demonstrating a significant boost in performance. 
This balance between data representation fidelity and model interpretability underscores the effectiveness of employing autograd-driven optimisation in fuzzy system~design. 

\begin{figure}[!ht]
    \begin{center}
        \includegraphics[trim=0 0 0 0, clip, width=.95\linewidth]{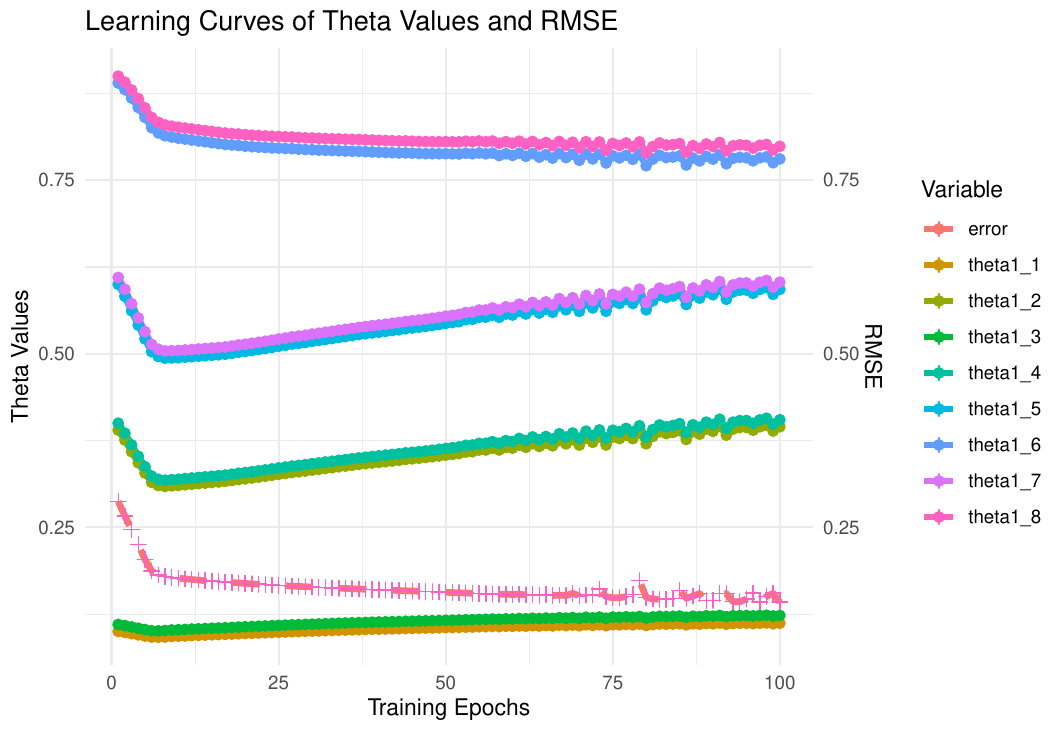}
        \caption{The learning curves of Parameters theta1 and RMSE}
        \label{fig:LearningCurves}
    \end{center}
\end{figure}

The learning curves of parameters \emph{theta1} and RMSE can be seen in Figure~\ref{fig:LearningCurves}.
The consistency of these improvements over multiple training iterations not only validates the effectiveness of incorporating autograd into FuzzyR, but also highlights its practical applicability in optimising fuzzy inference models. 

\section{Conclusion}
In this study, we have explored the use of automatic differentiation to simplify gradient-based optimisation processes of fuzzy systems.
Through a practical example in FuzzyR, we showcased how existing fuzzy inference system frameworks may be modified to incorporate the capabilities of an automatic differentiation tool, specifically Torch for R.
Although our demonstration only used a relatively simple Mamdani-type model on a small dataset, it effectively illustrated the potential of automatic differentiation to simplify the optimisation process, eliminating the need for intricate derivative computations.
More importantly, this automatic differentiation methodology is versatile and can be adapted for various forms of fuzzy system modelling, including type-2 or even more advanced systems, provided the inference mechanisms are mathematically differentiable.

Additionally, while this paper did not propose a new design paradigm for fuzzy systems, it highlights the transformative potential of automatic differentiation in the evolution of fuzzy system modelling.
By simplifying the complexity involved in computing derivatives, automatic differentiation may enable system designers to focus more on the innovation of system architectures. 
It could lead to a more straightforward exploration of a broader range of modelling configurations, such as a variety of membership functions and operators. 
This advancement could facilitate a shift toward more flexible and efficient design methodologies in the development of fuzzy~systems.

However, it is important to note that this study serves primarily as a demonstration purpose. 
There remains substantial scope for further research and development to fully exploit the capabilities of automatic differentiation in the design and optimisation of fuzzy systems. 
For example, considerable work may be needed to ensure that toolkits, such as FuzzyR, are fully compatible with automatic differentiation features. 
The ideal aim is to enhance fuzzy systems not only to retain their acclaimed advantages in explainability and interpretability but also to achieve or potentially exceed the performance benchmarks established by deep learning models.  
We hope that our work could inspire continued innovation and advancement in the field of fuzzy system modelling.


\bibliographystyle{IEEEtranNNoURL}
\bibliography{library}

\end{document}